\documentclass[journal,comsoc]{IEEEtran}
\usepackage[T1]{fontenc}
\usepackage[pdftex]{graphicx}
\usepackage[tight,footnotesize]{subfigure}

\ifCLASSINFOpdf
\else
\fi
\usepackage{amsmath}
\usepackage{txfonts}
\begin{document}
%
\title{Computational Models for Multiview Dense Depth Maps of Dynamic Scene}
%
%
%

\author{Qifei Wang
\thanks{Qifei Wang was with the Department of Electrical Engineering and Computer Sciences, University of California, Berkeley, CA 94720, USA, qifei.wang@gmail.com}
}

%
%

\markboth{Journal of \LaTeX\ Class Files,~Vol.~14, No.~8, August~2015}%
{Shell \MakeLowercase{\textit{et al.}}: Bare Demo of IEEEtran.cls for IEEE Communications Society Journals}
%



\maketitle



%
\IEEEpeerreviewmaketitle

\begin{abstract}
This paper reviews the recent progresses of the depth map generation for dynamic scene and its corresponding computational models. This paper mainly covers the homogeneous ambiguity models in depth sensing, resolution models in depth processing, and consistency models in depth optimization. We also summarize the future work in the depth map generation. 
\end{abstract}

\section{Introduction}
%
%
%
%
\IEEEPARstart{W}{ith} recent rapid advances in three-dimensional (3D) vision in both academia and industry, dense depth computation, especially the multiview depth for complex dynamic scene has become a new paradigm to the field of traditional computer vision. Obtaining dense multiview depth maps is the prerequisite for many challenging problems, such as dynamic scene modeling and reconstruction, 3D object recognition and tracking \cite{zhang2007multiview}, 3D reconstruction \cite{cao20123d}, 3D video coding \cite{wang2012complexity}\cite{wang2011reduced}\cite{wang2013complexity} and streaming \cite{ji2014online}\cite{wang2012free}\cite{wang2010region}, etc. 

In order to solve those challenges, some problems should be treated properly.

(1) Low temporal and interview stability: robust performance for vision-based researches and applications highly depends on the stability of multiview depth maps in temporal, spatial, and interview domains. However, this basic requirement often fails to be satisfied due to many reasons, where one of them is homogeneous ambiguity. These ambiguities are caused by homogeneous frequency among multiple time-of-flight (TOF) sensors or homogeneous structural light pattern among multiple RGB-D sensors, and these are treated as homogeneous ambiguity. 

(2) Resolution mismatch: depth maps record the 3D coordinates corresponding to those visible pixels in color image. However, since the spatial resolutions of depth sensors are usually lower than that of the CCD sensors, the mismatch between the pixels in depth maps and pixels in texture images may cause some errors in the processing steps using both texture images and depth maps. On the other hand, the temporal resolution mismatch between depth and CCD sensors is also serious problem in many 3D applications.

(3) Low precision: with high-precision depth information, the performance of some traditional vision problems with only texture images can be significantly improved. However, capability limitation \cite{wang2015evaluation} of depth sensor such as the noise level, limitations in dynamic cases for phase-based structure light system and others bring difficulties to obtain an accurate dense depth map. 

In this paper, we present the recent progresses in the field of dense depth computational models for dynamic scenes, especially the above criterion and difficulties.

\section{Stability models}
\begin{figure}[!htbp]
	\centering     
	\subfigure[Temporal interference in multiple TOF sensor systems. Part (A) is captured by dual TOF sensors, and (B) is by single.]{\label{fig:f}
		\includegraphics[trim=0 0 0 0, clip, width=0.4\textwidth]{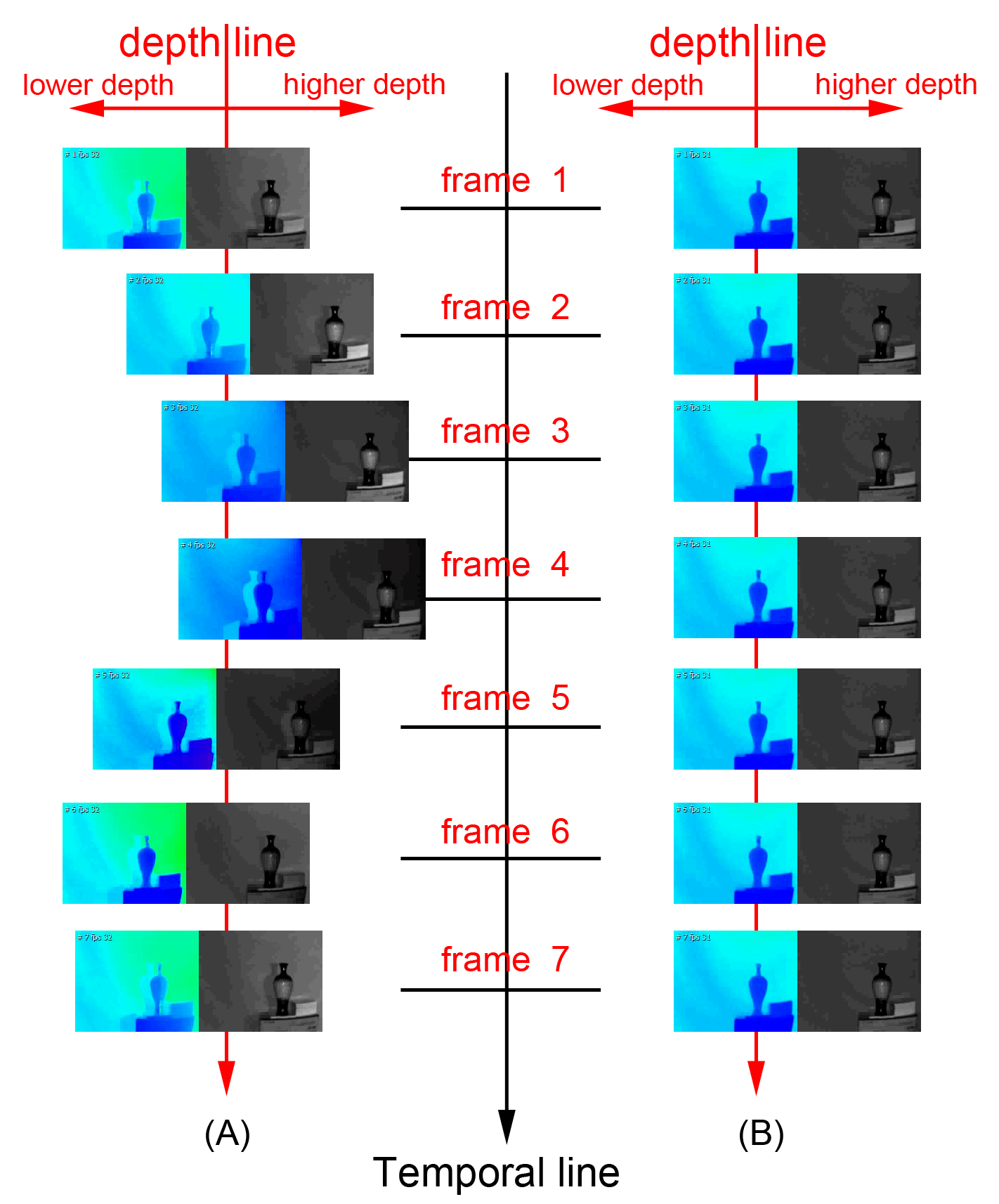}}
	\\
	\subfigure[Spatial interference in multiple RGB-D sensor system.Part (A) is captured by single RGB-D camera, and (B) is by dural carmera system \cite{maimone2012reducing}]{\label{fig:l}
		\includegraphics[trim=0 0 0 0, clip, width=0.4\textwidth]{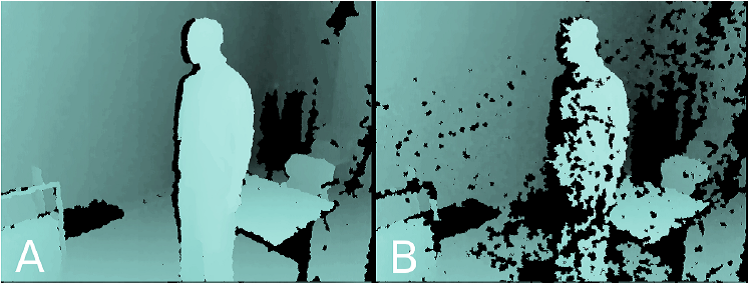}}
	\caption{Homogeneous ambiguities in multiple depth sensors.}
	\label{fig1:Homogeneous}
\end{figure}

The recent progresses on depth sensing, such as RGB-D and TOF sensors, have spurred developments of dynamic 3D scene modeling, markerless motion capture and 3D object motion tracking. However, multiple units with overlapping views cause prominent ambiguities, resulting in holes, noises, and interference in the computed depth maps. Examples of homogeneous ambiguities are given in Fig. \ref{fig1:Homogeneous} (a) and (b) which correspond to multiple TOF and RGB-D sensors, respectively. As can be found, the artifacts due to homogeneous ambiguity are not predictable in the captured depth maps when comparing to the results of single depth sensor.

\begin{figure}[!htbp]
	\centering 
	\includegraphics[trim=0 0 0 0, clip, width=0.4\textwidth] {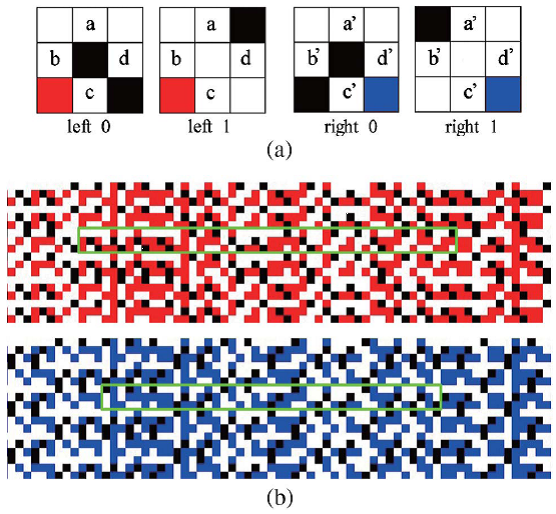}
	\caption{Binary codes and the hierarchical modules for spatial decoupling in multiple RGB-D depth sensing.}
	\label{fig2:Binary}
\end{figure}

It is a challenging problem to model the homogeneous ambiguity between these multiple depth sensors, because the homogeneous frequencies or the structural light pattern modules are coupled tightly. Decoupling models were proposed in spatial or temporal manners. In \cite{yan2014beyond}, a spatial decoupling method was proposed via hierarchical \emph{De Bruijn} binary modules. The method includes encoding and decoding stages, where hierarchical modules are set for encoding. Fig. \ref{fig2:Binary} shows an example for two coupled modules. Fig. \ref{fig2:Binary}(a) provides the binary codes for these two modules, and they are the lowest level in the hierarchical modules. After that, these binary codes are organized by rows and columns satisfying \emph{De Bruijn} rules, and therefore the hierarchical modules are given in Fig. \ref{fig2:Binary}(b).

The proposed model in \cite{yan2014beyond} is able to decouple the homogeneous ambiguity of multiple RGB-D sensors without degradation on depth map resolutions, and it brings great conveniences to 3D applications. Other than this model, other methods were proposed in temporal manners for multiple RGB-D sensors. For example, cyclic motion with different phrase was utilized to different RGB-D cameras, and each camera can capture sharp structural light pattern of its own while coupled patterns from other RGB-D cameras are blurred due to the induced relative motion \cite{maimone2012reducing}. Post-processing on blurred pixel separation is needed for this method. Besides that, a time-multiplex system was designed by a steerable hardware-shutter to simulate different cycles for corresponding RGB-D cameras \cite{berger2011markerless}. In this case, depth maps from different RGB-D cameras are in different time-slot, further temporal calibration is needed for temporal resolution up-conversion.

As for TOF sensors, so far, coprime frequencies are utilized in different sensors to avoid homogeneous ambiguities. Actually, the number of available frequency bands is up to 3 in many TOF cameras, and thus brings restrictions to multiview dense depth map capturing.

\section{Resolution models}
High spatial resolution of depth map is usually a result of stereo matching methods on high spatial resolution color image pairs rather than depth sensing. On the other hand, the temporal resolution depends on the sensing rate of depth sensors. For example, RGB-D camera can work with video rate (e.g. $30$ fps) and $640\times480$ resolution, while TOF camera can work with up to 60 fps but only $176\times144$ pixels. These capacities are far away from many practical usages.

\begin{figure}[!htbp]
	\centering     
	\subfigure[The cross-view depth up-sampling procedure.]{\label{fig:f}
		\includegraphics[trim=0 0 0 0, clip, width=0.4\textwidth]{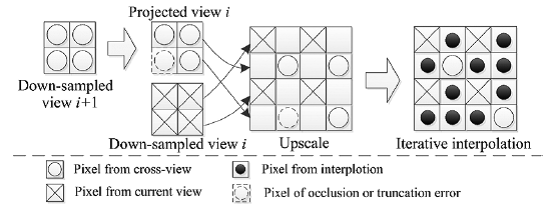}}
	\\
	\subfigure[The iterative filter for Upscale in (a).]{\label{fig:l}
		\includegraphics[trim=0 0 0 0, clip, width=0.4\textwidth]{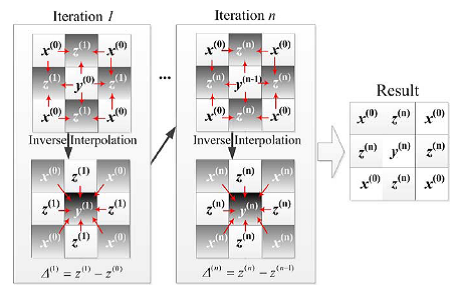}}
	\caption{Diagram of the cross-view depth up-sampling procedure for multiview depth.}
	\label{fig3:upsample}
\end{figure}

Iterative filter was proposed in \cite{liu2012cross} to up-sample the multiview depth maps simultaneously. This filter is originally proposed for multiview depth video coding to improve the rate-distortion performance. At the encoder side, multiview depth maps are down-sampled by odd-even interlaced extraction pattern. Then, the depth maps can be up-sampled via cross-view reference. The reference relationship is described in Fig. \ref{fig3:upsample}, and the iterative filter is modeled as

\begin{equation}\label{eq:1}
\begin{split}
& \mathbf{Z}^{(n)} = \mathbf{A}_{k\times m} \mathbf{X}^{(0)}+\mathbf{B}_{k\times l} \mathbf{Y}^{(n)} \\
& \mathbf{Y}^{(n)} = \mathbf{C}_{l\times m} \mathbf{X}^{(0)}+\mathbf{D}_{k\times l} \mathbf{Z}^{(n-1)} \\
\end{split}
\end{equation}

where $n$ is the number of iteration. $\mathbf{X}$ , $\mathbf{Y}$ and $\mathbf{Z}$ are pixel sets in Fig. \ref{fig3:upsample}(b). $\mathbf{A}$, $\mathbf{B}$ and $\mathbf{C}$ are coefficient matrix obtained through some selected up-sampling filters. The iterative filter is convergent and a unique result will be obtained.

Other than the iterative filters, some depth map super-resolution filters were proposed in two manners, which referring to or without referring to the corresponding view color image. As for color image assisted super-resolution, for example, filter parameters can be learned from corresponding the color image references \cite{yang2007spatial}\cite{ferstl2013image}\cite{yang2014stereo}. The sharp edges in depth up-sampling can also be reserved by aligning to the corresponding color image \cite{lo2013depth}. On the other hand, filter parameters can be fetched directly from low resolution depth maps \cite{ikehata2013depth}\cite{kim2012high}, or use joint bilateral union (JBU) filter simply and directly \cite{li2008hybrid}.

As for temporal resolution up-sampling, depth propagation is an effective approach. In this method, it is assumed that the variation for a given dynamic scene is identical for both the depth and color information of one viewpoint. Motion vector is widely utilized to describe the motion in dynamic scene, and propagation can be realized by this vector. The main problem for depth propagation is that it is very challenging to obtain accurate vectors for the occlusive or low textural regions. In order to solve the problem of motion vector accuracy, a rectification method was proposed in \cite{yang2012dynamic} by learning the surrounding features in color image. The decision in rectification is settled by a Heaviside step function

\begin{equation}
y=\lim_{k \rightarrow \infty} (\frac{1}{2}+\frac{1}{\pi}\tan^{-1}(kx))
\end{equation}

where $x\in (-\infty, +\infty)$. The rectification on motion vector improves the propagation performance significantly. In \cite{yang2012dynamic}, the experimental results showed that the temporal resolution can be eight times or even higher than the sensing rate in the proposed method.

\begin{figure}[!htbp]
	\centering 
	\includegraphics[trim=0 0 0 0, clip, width=0.4\textwidth] {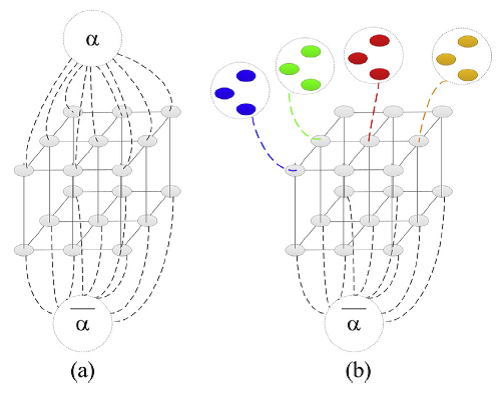}
	\caption{Graph models for energy minimization. Part (a) is the traditional model with one label set, and (b) is the multiple label set graph model.}
	\label{fig4:graphmodel}
\end{figure}

Recently, it was found that the up-conversion of temporal resolution for multiview depth computation is an energy minimization problem \cite{yang2014bundled}, and the traditional computational models can be utilized for this purpose. In \cite{yang2014bundled}, the computation of depth value is a selection from multiple candidate sets, and each of the elements of this set comes from different temporal and inter-view references. The element is selected by the motion vector. These candidates are with the same value range but different temporal or inter-view properties, i.e., different label set. Therefore, the best selection of depth value is an energy minimization of multiple label sets $E(L)$ rather than single label set in traditional model, and the model is as below

\begin{equation}
\begin{split}
\min E(L) &= E_{data}(L)+E_{smooth}(L) \\
& = \sum_{B_i \in I; l_x, l_y \in L_i} \Gamma_d(l_x, l_y) + \sum_{B_i \in I; l_x \in L_i, l_y \notin L_i} \Gamma_s(l_x, l_y) 
\end{split}
\end{equation}
where $L_i$ is the label sets, $l_x$ and $l_y$ are different labels. The subsequent computation is candidate selection via multi-set graph model, which is given by Fig. \ref{fig4:graphmodel}.

\section{Precision models}
Although depth sensors provide invaluable information for many 3D researches, their imaging capabilities are very limited in terms of noise level. The problem of consistency among depth maps in temporal, spatial and inter-view domains is still a challenge in depth map optimization.

The processing on depth map satisfies the criterion of Markov random field, and the problem of consistency can be modeled in stochastic field. In \cite{liu2013bayesian}, the inconsistencies was optimized by a risk function

\begin{equation}
\begin{split}
& \min_{\omega(1,2,\dots, n)} \sum_{i = 1,2,\dots,n}R(\omega(i)) \\
& s.t.\quad \omega(i) \in S
\end{split}
\end{equation}

where $\omega(.)$ is possible depth value, and $R(.)$ is the risk for selecting $\omega(.)$. Then, $R(.)$ can be obtained through Bayesian modeling as

\begin{equation}
R(\omega \mid f_i) = \sum_{l=0}^{2} \lambda(\omega, \omega^l) \frac{p(f_i \mid \omega)p(\omega)}{p(f_i)}
\end{equation}

where $f_i$ is a condition, $p(f_i \mid \omega)$ and $p(f_i)$ is conditional and prior probability, respectively. Then consequently, $p(f_i)$ can be learned through the initial depth maps, and the optimization model can be refined iteratively. The model has shown satisfied performance of multiview dense depth map optimization on temporal, spatial and inter-view consistency.

Other than the above models, traditional energy minimization model can also be utilized via proper defined smooth term in the model. For example, the gradient in both spatial and temporal domains are measured for corresponding domains consistency optimization.

The system capability of depth sensing can be improved significantly with the help of all above computation models. A hybrid camera system was built up in \cite{lee2011generation}, and the obtained dense depth maps for dynamic scene are selected by MPEG as standard test sequences.

\section{Conclusion}
In this paper we present recent progresses in depth computational models for dynamic scene, and the models cover the main processing chain in obtaining high quality dense depth maps. We discuss homogeneous ambiguity models in depth sensing, resolution models in depth processing, and consistency models in depth optimization. Although there is still a long way for high quality depth sensing, the mentioned models set up a new starting point to make further progresses.


%

%
%
%
%
%

\ifCLASSOPTIONcaptionsoff
  \newpage
\fi



%

%
%

\bibliographystyle{IEEEtran}
\bibliography{depth}

%

\begin{IEEEbiography}{Qifei Wang}
received the B.S. degree in information and computing science from Beijing University of Posts and Telecommunications, China, in 2007, and Ph.D degree in control science and engineering from Tsinghua University, China, in 2013. He joined EECS, University of California, Berkeley, US, in 2014. His current research interests include computer vision, machine learning, video processing and communications. 
\end{IEEEbiography}







\end{document}